\documentclass[10pt,twocolumn,letterpaper]{article}

\usepackage{cvpr}              
\usepackage[accsupp]{axessibility} 

\usepackage{multirow}
\usepackage[dvipsnames]{xcolor}

\newcommand{\textsmaller}[1]{\scalebox{0.8}{#1}}
\newcommand{\ours}{\texttt{%
    \textbf{%
        L\textsmaller{A}V\textsmaller{I}T\textsmaller{I}%
    }%
}%
}%

\usepackage{amsmath,amsfonts,bm}
\usepackage{xspace}
\makeatletter
\DeclareRobustCommand\onedot{\futurelet\@let@token\@onedot}
\def\@onedot{\ifx\@let@token.\else.\null\fi\xspace}

\def\eg{\emph{e.g}\onedot}

\makeatother

\def\eqref#1{equation~\ref{#1}}

\def\1{\bm{1}}

\def\rl{{\textnormal{l}}}

\def\rv{{\textnormal{v}}}

\def\rmL{{\mathbf{L}}}
\def\rmM{{\mathbf{M}}}

\def\rmQ{{\mathbf{Q}}}

\def\rmT{{\mathbf{T}}}

\def\rmV{{\mathbf{V}}}

\def\rmY{{\mathbf{Y}}}

\DeclareMathAlphabet{\mathsfit}{\encodingdefault}{\sfdefault}{m}{sl}
\SetMathAlphabet{\mathsfit}{bold}{\encodingdefault}{\sfdefault}{bx}{n}

\newcommand{\R}{\mathbb{R}}

\definecolor{cvprblue}{rgb}{0.21,0.49,0.74}
\usepackage[pagebackref,breaklinks,colorlinks,citecolor=cvprblue]{hyperref}

\def\paperID{16} 
\def\confName{CVPR}
\def\confYear{2024}

\title{Contrastive Language Video Time Pre-training}

\author{Hengyue Liu$^{1,2}$ \thanks{Work done during an internship at Intel Labs.}
\quad Kyle Min$^{2}$ \quad Hector A. Valdez$^{2}$ \quad Subarna Tripathi$^{2}$\\
$^{1}$UC Riverside \quad\quad $^{2}$Intel Labs\\
{\tt\small hliu087@ucr.edu, \{kyle.min, hector.a.valdez, subarna.tripathi\}@intel.com}
}

\begin{document}
\maketitle

\begin{abstract}
We introduce \ours, a novel approach to learning language, video, and temporal representations in long-form videos via contrastive learning.
Different from pre-training on video-text pairs like EgoVLP, \ours\ aims to align language, video, and temporal features by extracting meaningful moments in untrimmed videos.
Our model employs a set of learnable moment queries to decode clip-level visual, language, and temporal features.
In addition to vision and language alignment, we introduce relative temporal embeddings (TE) to represent timestamps in videos, which enables contrastive learning of time.
Significantly different from traditional approaches, the prediction of a particular timestamp is transformed by computing the similarity score between the predicted TE and all TEs.
Furthermore, existing approaches for video understanding are mainly designed for short videos due to high computational complexity and memory footprint.
Our method can be trained on the Ego4D dataset with only 8 NVIDIA RTX-3090 GPUs in a day.
We validated our method on CharadesEgo action recognition, achieving state-of-the-art results.

\end{abstract}    
\section{Introduction}
\label{sec:intro}

In recent years, there has been a surge
of interest in developing egocentric video understanding models leveraging 
video-text pre-training, followed by finetuning
for downstream applications. 
A line of work \citep{Lin2022c,Pramanick2023o,Zhao2022l} aiming to learn transferable spatio-temporal features from large video-text datasets have been emerged. Methods such as LAVILA \citep{Zhao2022l} showed how leveraging the dense narrations generated by Large Language Models (LLM) can be beneficial for video-language pre-training. However, all such methods hit the memory and compute-bottleneck while processing video sequences each with a few number of frames, leading to the reasoning capacity of the video models in a limited temporal context. Additionally, the above models do not use explicit temporal reasoning. 
In this work, we propose \ours, aiming to align language, video, and temporal features by extracting meaningful moments in untrimmed videos and equip the model with long-form temporal reasoning capability in an memory and compute efficient way. 
\ours{} can be evaluated on zero-shot episodic memory tasks such as natural language query (NLQ), thanks to the integration of explicit temporal modeling over \emph{untrimmed} videos into the pre-training objective.

The key contributions of this work are:
(1) aligning
language, video, and \emph{temporal features} by extracting meaningful moments in \emph{untrimmed} videos;
(2) formulating the video, language and temporal alignment as a direct set prediction problem;
(3) 
enabling long-form reasoning over potentially thousands of frames of a video in a memory-compute efficient way;
(4) demonstrating the efficacy of \ours{} by its superior performance on CharadesEgo action recognition;
{(5) Enabling zero-shot natural language query (NLQ) task without needing to train additional sub-networks or NLQ annotations.

\section{Related Work}
\label{sec:related_work}


In recent years, egocentric video-language pre-training (VLP) has been adopted significantly in academia and in industry.
A line of works 
such as EgoVLP \citep{Lin2022c}, EgoVLPv2 \citep{Pramanick2023o} learn transferable 
spatial-temporal representation from large-scale video-text datasets. Recently, LaViLa \citep{Zhao2022l} showed that VLP can benefit from the dense narrations
generated by Large Language Models (LLMs).
However, all such methods do hit the memory and compute bottleneck while processing video sequences, each consisting of a small number of frames (\eg 8 or 16 frame models), leading to limited temporal context aggregation capability. 
On the contrary, \ours\ is equipped with long-form reasoning capability (1,000 frames vs 16 frames) and is not limited to a small number of input frames from a video sequence.
\section{Approach}

The primary goals of our pre-training method are: (1) 
capturing 
temporal dynamics of videos, 
(2) 
aligning 
language, visual, and temporal information at clip level, and (3) 
ability to 
efficiently 
process 
long-form
videos.
We use frozen CLIP~\cite{Radford2021u,IlharcoGabrielAndWortsmanMitchellAndWightmanRossAndGordonCadeAndCarliniNicholasAndTaoriRohanAndDaveAchalAndShankarVaishaalAndNamkoongHongseokAndMillerJohnAndHajishirziHannanehAndFarhadiAliAndSchmidtLudwigOtherd} vision and text encoders for feature extractions. Temporal modeling is performed over the extracted CLIP visual features and learnable relative temporal embeddings via a transformer~\cite{Vaswani2017s} encoder.
We then create a set of learnable moment queries to directly predict the language and temporal embeddings of moments in videos through a transformer decoder.
The overall architecture and training of \ours\ is illustrated in Fig.~\ref{fig:arch}.

\begin{figure}
    \centering
    \includegraphics[width=\linewidth]{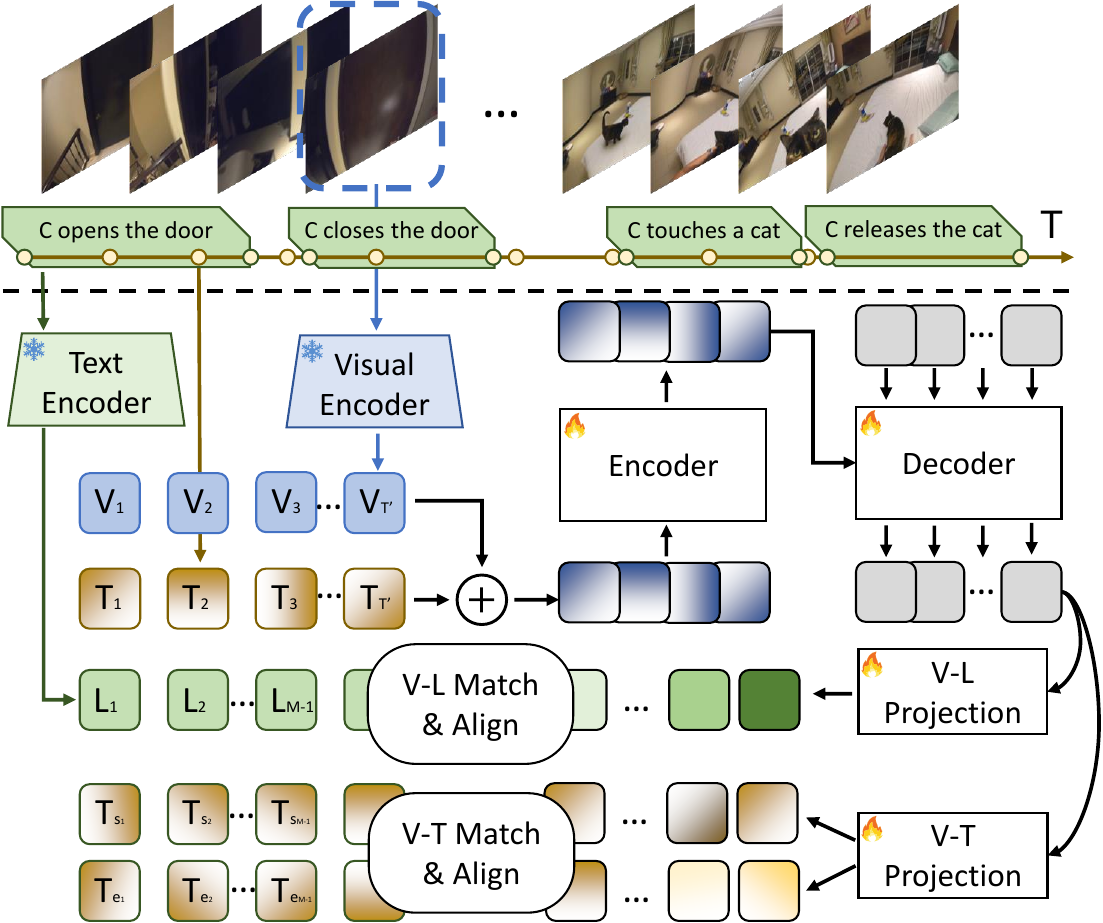}
    \caption{The architecture and training pipeline of \ours{}. We use a set of learnable queries to capture both visual and temporal features, and directly predict the visual (V) and temporal (T) embeddings of potential moments, respectively. Predicted visual embeddings are aligned with ground-truth narration text embeddings (L), and predicted TE are aligned with interpolated TE at ground-truth timestamps.}
    \label{fig:arch}
\end{figure}

\subsection{Feature Extraction}

Different from spatio-temporal transformers~\cite{Bertasius2021m}, which can only utilize a fixed number of frames, our method adopts a post-temporal information injection strategy to enable long-form temporal reasoning.
We use the CLIP vision encoder, \eg, ViT~\cite{Dosovitskiy2020e}, to extract the \texttt{[cls]} token from each frame independently.
The visual features of a video can be represented as a sequence of visual vectors $ \rmV = \{ \rv_1, \rv_2, \dots, \rv_T \} $, $ \rv_t \in \R^{1 \times C} $, where $T$ is the number of frames in the video, and $C$ is the channel dimension of the frame features.
We then save the extracted frame features to storage, so that we can perform off-line pre-training without accessing video data.
This significantly reduces the computation burden, and we can train on whole untrimmed videos rather than short-term video sequences. 
We then employ a non-overlap 1D convolution on $\rmV$ to generate $T'$ number of visual tokens with a feature dimension of $d$, denoted by $\rmV' \in \R^{T' \times d}$.
For the pre-training purpose, the frozen text encoder is used to create the language embeddings of all narrations by extracting the $\texttt{[eos]}$ tokens.
Given $M$ number of narrations in a video, we extract a set of language vectors $ \rmL = [ \rl_1, \rl_2, \dots, \rl_M ] $, $ \rl_j \in \R^{1 \times C} $, and each vector is L2-normalized.

\subsection{Temporal Embedding and Encoder}

To inject temporal information, we create learnable temporal embeddings $ \rmT \in \R^{T_0 \times d} $, representing all the timestamps of a video.
$\rmT$ is initialized as 1D positional embeddings following~\cite{Vaswani2017s}, where the cosine similarity between adjacent embeddings is larger distant ones.   
As $T' \neq T_0$, we interpolate $ \rmT $ to the same length of $T'$ as $\rmT'$, where each embedding can represent a particular timestamp relatively to the video length.
The visual tokens $\rmV'$ are added with temporal embeddings $ \rmT' $ as the video features $ \rmM = \rmV' + \rmT' $, which is fed into a standard transformer encoder.

\subsection{Decoder}

Instead of aligning a video clip $\texttt{[cls]}$ token with the corresponding text embeddings of the description of the clip (\eg, a narration), we view video-language pre-training on untrimmed videos as a direct set prediction problem.
Following DETR~\cite{Carion2020d}, we create fixed-size learnable queries $\rmQ \in \R^{N \times d}$, where the video features $\rmM$ serve as keys and values to the decoder.
Denote the output embeddings of the decoder as $\rmQ'$, they are projected and L2-normalized into visual embeddings $\rmV' \in \R^{N \times C}$ and temporal embeddings $\rmT' \in \R^{N \times 2 \times d}$ via feed-forward networks (FFNs), respectively.
The output of the decoder is denoted by $\rmY' = \{(\rmV'_i, \rmT'_{s_i}, \rmT'_{e_i})\}_{i=1}^N $, where temporal embeddings corresponded to the start and end timestamps $(s_i, e_i)$ of detected moments.

\subsection{Match and Alignment}

Since the narrations in the Ego4D dataset are annotated with a single timestamp rather than an interval, we augment each narration with a start and end timestamp.
Different from EgoVLP, a narration's start (or end) timestamp is determined by its previous and later narrations.
For a narration with timestamp $t_j$, we uniformly sample a start and end timestamp $(s_j, e_j)$ as

\begin{equation}
    \begin{aligned}
        &s_j = \texttt{Uniform}(t_{j-1}, t_{j})\\
        &e_j = \texttt{Uniform}(t_{j}, t_{j+1}).
    \end{aligned}
\end{equation}
For each narration, we can sample its corresponding temporal embeddings $ \rmT_{s_j} $ and $ \rmT_{e_j} $.
Let the ground truth moments be $\rmY = \{(\rmL_j, \rmT_{s_j}, \rmT_{e_j})\}_{j=1}^M $, we perform the bipartite matching between $\rmY$ and $\rmY'$ via Hungarian algorithm.
Concretely, we compute 3 pairwise cosine similarities $<\{\rmV'_i\}_{i=1}^N, \{\rmL_j\}_{j=1}^M>$, $<\{\rmT'_{s_i}\}_{i=1}^N, \{\rmT_{s_j}\}_{j=1}^M>$, and $<\{\rmL'_{e_i}\}_{i=1}^N, \{\rmL_{e_j}\}_{j=1}^M>$.
Each similarity matrix is followed by the Sigmoid activation, which allows for multi-label matching as there exist the same narrations or timestamps within videos.
The final cost is the negation of the element-wise product of the 3 similarity matrices.
After finding the matched predictions and groundtruth, following SigLIP~\cite{Zhai2023t}, we use Sigmoid contrastive loss to align language, vision and time of moments.
For unmatched predictions, we push their similarities with groundtruth to be -1 (equivalent to all labels of 0 in binary cross-entropy).


\section{Experiments}
\label{sec:results}

\subsection{Implementation Details}
We use the Ego4D~\cite{Grauman2022l} dataset for pre-training.
Each untrimmed video is divided into chunks of 600 seconds following~\cite{Lin2022c} for efficient data access.
Our codebase is adopted from OpenCLIP~\cite{IlharcoGabrielAndWortsmanMitchellAndWightmanRossAndGordonCadeAndCarliniNicholasAndTaoriRohanAndDaveAchalAndShankarVaishaalAndNamkoongHongseokAndMillerJohnAndHajishirziHannanehAndFarhadiAliAndSchmidtLudwigOtherd} and LAVILA~\cite{Zhao2022l}.
We use the CLIP vision encoder with ViT-H-14 backbone pre-trained on DFN-5B~\cite{Fang2023g}, and standard CLIP text encoder~\cite{Radford2021u}.
The visual embeddings are pre-extracted and stored locally with a stride of 5, namely we uniformly sample and compute the embeddings of 6 frames per second.
We train the model with 8 NVIDIA RTX-3090 GPUs for 20 epochs with a batch size of 256, and a learning rate of $5\times10^{-4}$ using the Adam optimizer~\cite{Kingma2014j}.
It takes approximately 20 minutes to train 1 epoch.
The 1D convolution layer has a kernel size of 7 with a stride of 7, and the output number of channels $d=512$.
Both the transformer encoder and decoder has a stack of 6 layers, with 8 attention heads and each head has a feature dimension of 64.

\subsection{Action Recognition}

We evaluate our method on one of the downstream tasks, namely action recognition. 
We use the CharadesEgo~\cite{Sigurdsson2018q} dataset under both the zero-shot (ZS) and finetuning (FT) settings.
We report video-level mAP as the evaluation metric following previous works~\cite{Lin2022c,Zhao2022l}. 
As our method is capable of processing long-form videos, we use the whole video for training and testing without the need of sampling~\cite{Lin2022c,Zhao2022l}.
For this task, we use the averaged similarity scores between all output embeddings $\{\rmL'_i\}_{i=1}^N$ with each ground-truth label.

The results are shown in Table~\ref{tab:charadesego}.
In both settings, \ours{} outperforms all video foundation models by a large margin.
\ours\ outperforms GPT4Ego-L and LAVILA-L under the ZS setting by 3.0 and 5.6, respectively.
\ours\ also achieves 1.9 improvement over LAVILA-L under the FT setting.
Comparing with other methods, we can perform prediction with arbitrary number of frames instead of fixed number of frames. 
It is also worth noting that both GPT4Ego and LAVILA use LLMs for either training or testing to augment the language representations, whereas we use a frozen text encoder.

\begin{table}[ht]
  \centering
  \begin{tabular}{@{}l c c c @{}}
    \toprule
    Method & Backbone & mAP (ZS) & mAP (FT) \\
    \midrule
    EgoVLP~\cite{Lin2022c} & TSF-B & 25.0 & 32.1 \\
    EgoVLPv2~\cite{Pramanick2023o} & TSF-B & 26.2 & 34.1 \\
    LAVILA~\cite{Zhao2022l} & TSF-B & 26.8 & 33.7 \\
    LAVILA~\cite{Zhao2022l} & TSF-L & 28.9 & \underline{36.1} \\
    GPT4Ego~\cite{Dai2024o} & TSF-B & 29.6 & - \\
    GPT4Ego~\cite{Dai2024o} & TSF-L & \underline{31.5} & - \\
    \midrule
    \ours & ViT-H-14 & \textbf{34.5} & \textbf{38.0} \\
    \bottomrule
  \end{tabular}
  \caption{Performance on CharadesEgo. \ours\ achieves significant gains in both zero-shot and fine-tuned settings.
  ZS and FT stand for zero-shot and finetuning, respectively.}
  \label{tab:charadesego}
\end{table}

\subsection{Natural Language Query}

As \ours{} is capable of long-form video understanding with explicit temporal alignment, the Ego4D Natural Language Query (NLQ) task is a natural fit with the pre-training targets.
We can directly predict intervals which are aligned with language query given a video; therefore, \ours{} can perform the NLQ task under the zero-shot setting (without modifications of the architecture and re-training on NLQ annotations).
We can directly use the text embedding of the question to match with the predicted visual embeddings $\{\rmV'_i\}_{i=1}^N$, and select the top-K predicted temporal embeddings $\{(\rmT'_{s_i}, \rmT'_{e_i} )\}_{i=1}^K$ as the response to the question.
We then take the argmax indices of the similarities of the predicted temporal embeddings with $\rmT$, which can be mapped to timestamps.
We follow the standard evaluation metrics on NLQ, and report the recall@\{1, 5\} with IoU$\in$\{0.3, 0.5\}.
The preliminary ZS results are list in Table~\ref{tab:nlq_zs}.
\begin{table}[ht]
    \centering
    \begin{tabular}{ccccc}
        \toprule
        & \multicolumn{2}{c}{IoU=0.3} & \multicolumn{2}{c}{IoU=0.5} \\
        & R@1 & R@5 & R@1 & R@5 \\
        \hline
        \ours{} & 2.50 & 8.52 & 1.08 & 3.36 \\
        \bottomrule
    \end{tabular}
    \caption{ZS recall on the validation set of Ego4D NLQ benchmark.}
    \label{tab:nlq_zs}
\end{table}

In the near future, we plan on assessing its potential to learn improved representations for episodic memory tasks including NLQ and Moment Query (MQ).
\section{Conclusions}
\label{sec:conclusion}

We devise a novel approach to learning language, video, and temporal representations in long-form videos via contrastive learning, termed as \ours. Unlike existing methods, \ours\ aims to align language, video, and temporal features by extracting meaningful moments in untrimmed videos by formulating it as a direct set prediction problem.
Our method outperforms existing state-of-the-art methods by a significant margin on egocentric action recognition, yet is trainable on memory and compute-bound systems.

{
    \small
    \bibliographystyle{ieeenat_fullname}
    \bibliography{paperpile.bib}
}


\end{document}